\pdfoutput=1
\documentclass{article}

\usepackage[final]{nejlt}
\fancypagestyle{title}{\fancyhf{}\fancyfoot[C]{\small Preprint}\fancyfoot[R]{\small\thepage}%
  }
\usepackage{booktabs}
\usepackage{graphicx}
\usepackage{amsmath}
\usepackage{array}
\usepackage{multirow}
\usepackage{xspace}

\newcommand{\pyth}{Pythia\xspace}
\newcommand{\mamba}{Mamba\xspace}

\newcolumntype{L}[1]{>{\raggedright\arraybackslash}p{#1}}
\title{From Architecture to Output: Testing the Structural Origins of Hallucination and the Amplifying Role of Data}

\makeatletter
\edef\headtitle{From Architecture to Output}%
\makeatother

\author{Md.\ Rejaul Korim Sadi, Toufiqur Rahman Tasin, Golam Mostofa Naeem\\
Metropolitan University, Sylhet, Bangladesh\\
\texttt{rejaulkorimsadi@gmail.com}}

\begin{document}
\thispagestyle{title}

\abstract{Surveys of hallucination classify errors by how an output relates to its source, and often explain them by the model's structure: attention retrieves by association, the maximum-likelihood objective has no term for truth, and left-to-right decoding commits to early tokens. Data, on this account, makes these failures worse rather than causing them. The account is usually argued rather than tested. We test it with interventions that need only sampling access. Study~1 holds pretraining data, tokenizer and scale fixed and swaps attention for a state-space mixer, using five Pythia/Mamba pairs from 130M to 2.8B parameters and four further attention models, on 111 facts presented with a planted false or a planted true statement. Swapping the mixer leaves free-recall failure unchanged (both 2.8B models answer 3 of 200 SimpleQA questions), but it changes what happens to a defect in the input. From 410M upwards, every attention model repeats the planted falsehood more often than its matched state-space model (2.8B: 650 vs.\ 566 of 666 items, $p=2.5\times10^{-20}$), also on facts both models know, while the state-space models can still read planted facts they did not know. Study~2 replaces the first wrong fact in 111 erroneous answers from three instruction-tuned models. The correct fact cuts the share of answers with a downstream error from 92.5\% to 48.1\%, but a different wrong fact works as well ($p=0.38$), and the models give the right answer in isolation on only 2 of 111 items. The structural account survives in a narrower form. The mixing component decides whether a defect in the context becomes an output error, commitment decides how an error spreads, and on long-tail questions the first error comes from missing knowledge, not from a suppressed truth.}

\maketitle

\section{Introduction}

Language models state false things fluently and with confidence. The field describes these failures carefully. An error is \emph{intrinsic} when it contradicts the source and \emph{extrinsic} when the source cannot verify it \cite{maynez2020,ji2023}; for large language models (LLMs) the same split is drawn between faithfulness and factuality \cite{huang2023,alansari2025}. These labels describe a relation between an output and a reference. They say nothing about the computation that produced the output, and so nothing about where to intervene.

Explanations of \emph{why} models hallucinate tend to point at the architecture. Three parts of a decoder-only model come up again and again. Self-attention \cite{vaswani2017} retrieves whatever the context makes associatively available and has no operation that checks whether the retrieved content is valid in the present case. The maximum-likelihood objective rewards agreement with the corpus and contains no term for truth, so a frequent falsehood can outrank a rare fact \cite{lin2022,kalai2025}. Left-to-right decoding commits to every token it emits, and because training conditions on reference text while generation conditions on the model's own output, an early error is carried forward rather than revised \cite{ranzato2016,wang2020exposure,zhang2024snowball}. Data enters these accounts as an amplifier: long-tail sparsity \cite{kandpal2023,mallen2023}, skewed corpora and recycled synthetic text \cite{shumailov2024} give the components worse material to work with, but do not create the failure modes themselves.

We call this the \emph{structural account}. It is plausible and widely repeated, and it has rarely been put to an experimental test. It also tends to be stated too strongly. ``Architecture matters more than data'' is not something any current evidence shows, and ``every data-induced error passes through a component'' cannot fail, since every output passes through every component. The question we ask is narrower and testable: if one component is changed while training data is held fixed, do the errors attributed to that component change, and do the others stay put?

Recent work on the behavioural evaluation of language models frames how far such a test can go. \newcite{callegari2026} argue that behavioural tests of LLMs show what a model does, not how it does it, and that performance-level evidence should not be read as evidence about mechanism. \newcite{pikuliak2024} reaches a similar conclusion for self-report studies. \newcite{vanmiltenburg2023error} find that few NLG papers analyse their system errors at all. We take these warnings at face value. Our interventions are behavioural, and we use them to attribute errors to components only where a control rules out the obvious alternative.

We run two studies. \textbf{Study~1} (Section~\ref{sec:study1}) tests the attention part of the account by swapping the context mixer at matched data and scale: five pairs of Pythia \cite{biderman2023pythia} and Mamba \cite{gu2023mamba} models trained on the same corpus with the same tokenizer, a probe of 111 facts with planted false and true statements, and controls for prior knowledge and for the ability to read the planted line. The planted falsehood doubles as a controlled data defect of the kind retrieval-augmented systems meet in practice. \textbf{Study~2} (Section~\ref{sec:study2}) tests the commitment part by turning the snowballing design of \newcite{zhang2024snowball} into an intervention with two controls, on three instruction-tuned models including a reasoning model.

The results split. In Study~1 the attention attribution survives every control we apply: at matched data, whether a planted falsehood becomes an output error depends strongly on the mixer. In Study~2 the commitment effect is large, but its usual explanation, that the model knew the truth and was trapped by its own earlier token, does not hold on these items.

Our contributions are:
(i) a sampling-only version of the structural account, with one intervention and one falsifiable prediction per component and a testable form of the claim that data amplifies rather than originates;
(ii) a matched-architecture test at five scales in which removing attention leaves free-recall failure unchanged but reduces adoption of misleading context, with reading ability and prior knowledge controlled, and an error analysis by fact type, text length and position;
(iii) a controlled prefix-repair experiment that separates the behavioural effect of commitment from its knowledge-based explanation;
(iv) a discussion of what the results imply for instruction tuning, preference optimisation, retrieval, scale and calibration.
Raw outputs, blind scoring sheets and analysis code are released with this preprint as ancillary files.

\section{Related Work}
\label{sec:related}

\paragraph{Output taxonomies.} The intrinsic/extrinsic split was introduced for summarisation \cite{maynez2020}, extended across generation tasks by \newcite{ji2023} and recast for LLMs as faithfulness versus factuality \cite{huang2023}. \newcite{alansari2025} add finer types such as factual contradiction, entity error and logical inconsistency. None of these schemes locates an error inside the model, and none was designed to. Section~\ref{sec:framework} adds a second, component axis to them.

\paragraph{Behavioural analysis.} \newcite{callegari2026} and \newcite{pikuliak2024} caution against reading mechanism or competence from behaviour. \newcite{dayanik2022} attribute differences in model output to specific factors with regression, covariates and effect sizes, controlling confounds instead of testing one factor at a time. \newcite{vanmiltenburg2023error} call for systematic error analysis in NLG, and \newcite{vanmiltenburg2023registered} for fixing the analysis plan before data collection; Study~2 follows that practice and reports its deviations.

\paragraph{Data and frequency.} \newcite{dziri2022} find that conversational models amplify hallucination already present in their training data. \newcite{mckenna2023} trace entailment errors to attestation and relative-frequency biases. Accuracy falls with how rarely an entity appears in pretraining data \cite{kandpal2023,mallen2023}. \newcite{kalai2025} argue that hallucination follows from pretraining statistics and persists because evaluations reward guessing over abstaining. Training on model-generated text narrows the learned distribution and erodes its tails \cite{shumailov2024}.

\paragraph{Knowledge conflicts and context faithfulness.} When the context contradicts what a model learned in pretraining, models often side with the context, to a degree that varies with model and evidence \cite{longpre2021,xie2024,wu2024clasheval}; \newcite{xu2024conflicts} survey this literature. Models are also distracted by irrelevant context \cite{shi2023distracted} and use evidence unevenly by position \cite{liu2024}, with attention mass gathering on early tokens regardless of content \cite{xiao2024sinks}. Retrieval-augmented generation \cite{lewis2020} inherits these weaknesses, and a few injected passages can steer its answers \cite{zou2025poisonedrag}. Preference-trained assistants show a related deference to the user's stated view, known as sycophancy \cite{sharma2024sycophancy}.

\paragraph{Mechanistic localisation.} Factual associations are stored largely in feed-forward layers \cite{geva2021,meng2022}, and attention moves attributes to the prediction position \cite{geva2023}. Induction heads implement in-context copying \cite{olsson2022}. State-space models are weaker than transformers at associative recall and copying from context \cite{arora2024zoology,jelassi2024repeat}, which bears directly on Study~1 (Section~\ref{sec:s1-limits}). Internal directions can be read out and used to steer generation \cite{li2023iti,leino2025}, and internal states often encode whether an answer is correct when the output does not show it \cite{azaria2023,orgad2025,farquhar2024}. These methods need weights or activations; ours need only inputs and prefixes.

\paragraph{Exposure bias and snowballing.} Training on reference prefixes and generating from the model's own output creates exposure bias \cite{ranzato2016,bengio2015}, which \newcite{wang2020exposure} connect to hallucination in translation. Models partially recover from injected errors \cite{he2021exposure}. \newcite{zhang2024snowball} show that models defend an early wrong answer with further claims they reject when asked separately. Study~2 builds on their design.

\section{The Structural Account and Its Tests}
\label{sec:framework}

\paragraph{Three components.} We consider three components of a decoder-only model and the failure each is said to produce.
\emph{Context mixing} combines information across positions. In a transformer it is self-attention, which returns whatever the learned query--key geometry makes retrievable from the context; nothing in the computation asks whether the retrieved association holds for the case at hand. We call this \emph{associative retrieval without a validity check}.
The \emph{pretraining objective} minimises $-\sum_t \log P_\theta(x_t \mid x_{<t})$. Every term rewards agreement with the corpus and none refers to truth, so the objective yields \emph{fluency without a truth term}.
\emph{Decoding} is left to right: each token, once emitted, conditions everything after it, and since training conditions on reference prefixes $x^{*}_{<t}$ while generation conditions on the model's own $\hat{x}_{<t}$, the two distributions differ. The resulting failure is \emph{commitment under exposure bias}.
Existing models already vary each component with the other two fixed: state-space models change the mixer and keep the objective and decoding; preference optimisation \cite{ouyang2022,rafailov2023} changes the objective; decoding rules change without retraining. The list is not exhaustive. Feed-forward memory \cite{geva2021}, tokenisation and positional encoding are other candidates.

\paragraph{Data as amplifier, stated so that it can fail.} The strong version of the structural account holds that a data defect becomes an output error only by passing through a component. As stated, that cannot be tested. A weaker version can: \emph{at fixed data, the rate at which a given defect becomes an output error depends on the component it passes through}. If a falsehood placed in the context is repeated equally often by models with different mixers, the mixer adds nothing and the defect alone explains the error. If the rate differs, the component is doing part of the work. We test this version for two kinds of defect: misleading text in the context window (Study~1), which stands in for a bad retrieved passage or polluted web text, and sparse coverage of a fact in pretraining (Studies~1 and 2).

\begin{figure*}[t]
\centering
\includegraphics[width=\textwidth]{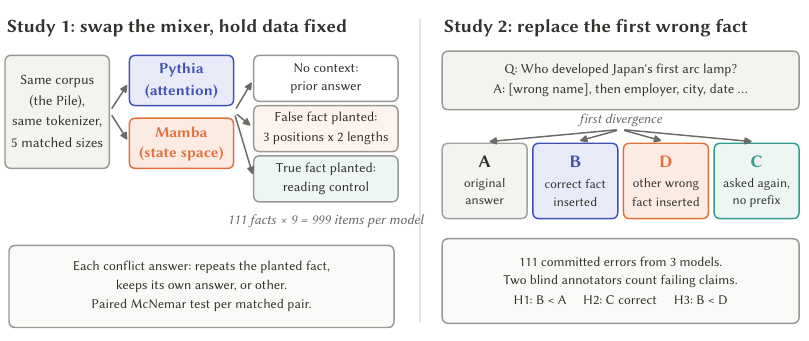}
\caption{The two studies. Study~1 (left) compares models that share training data and tokenizer but differ in how they mix context, on the same 999 items. Study~2 (right) takes answers whose first wrong fact was followed by further claims, rewrites that fact in two ways, and asks the question once more on its own.}
\label{fig:design}
\end{figure*}

\paragraph{Interventions.} For a prompt and an erroneous response, \textbf{T1 (prefix)} replaces the first divergent span with the correct fact and regenerates; as a control it replaces the span with a different wrong fact, and separately it asks for the fact with no prior commitment. \textbf{T2 (context)} varies the presence, truth and position of conflicting material in the context while holding the question fixed. \textbf{T3 (frequency)} reduces the prompt to a minimal query and tests whether the error matches the higher-frequency competitor across paraphrases. All three need only prompt and prefix control. Table~\ref{tab:framework} lists the component each one targets, the output type that component is expected to produce most often, and what this paper tests; Figure~\ref{fig:design} shows the two study designs.

\paragraph{Predictions tested here.}
\textbf{P5 (component substitution).} Replacing attention with a non-attention mixer at matched data and scale changes context-conflict errors and leaves free-recall errors roughly unchanged. It fails if neither changes or if both change together.
\textbf{P1 (position).} At fixed content, moving the conflicting line changes the error rate.
\textbf{P3 (commitment).} Correct-fact substitution at the first divergence reduces downstream errors. The knowledge-based version holds that this works because the model knows the right answer and the substitution frees it from an early commitment. It adds two predictions: the correct fact outperforms a wrong one (P3a), and the model answers correctly in isolation (P3b).

\begin{table*}[t]
\centering
\small
\renewcommand{\arraystretch}{1.15}
\setlength{\tabcolsep}{4pt}
\begin{tabular}{@{}L{2.3cm}L{2.9cm}L{2.4cm}L{2.4cm}L{2.7cm}L{2.8cm}@{}}
\toprule
Component & Proposed failure & Modal output type & Data amplifier & Test (this paper) & Outcome \\
\midrule
Context mixing (attention) & Associative retrieval without a validity check & Intrinsic; entity and context errors & Misleading or polluted context & T2 with matched mixer swap, P5 and P1 (Study~1) & Supported at four of five scales, with controls \\
\addlinespace
Pretraining objective (MLE) & Fluency without a truth term & Extrinsic; factual errors & Long-tail sparsity, skew & Free recall at matched data (Probe~1a); isolated query (Study~2) & Recall failure is mixer-independent; errors track missing knowledge \\
\addlinespace
Decoding (left to right) & Commitment under exposure bias & Logical inconsistency; error cascades & Errors in earlier output & T1 with wrong-fact and isolation controls, P3 (Study~2) & Behavioural effect holds; knowledge-based version fails \\
\bottomrule
\end{tabular}
\caption{The structural account as tested here. ``Modal output type'' is the type in the taxonomy of \protect\newcite{alansari2025} that each component is expected to produce most often; the mapping is a tendency, not a partition. The frequency test T3 is not run.}
\label{tab:framework}
\end{table*}

\section{Study 1: Swapping the Mixer at Matched Data and Scale}
\label{sec:study1}

\subsection{Setup}

\paragraph{Models.} Pythia \cite{biderman2023pythia} uses attention; Mamba \cite{gu2023mamba} uses a selective state space. Both families were trained on the Pile \cite{gao2020pile} with the GPT-NeoX tokenizer, which lets us form five pairs matched on data, tokenizer and approximate size: \pyth-160M/\mamba-130M, \pyth-410M/\mamba-370M, \pyth-1B/\mamba-790M, and the 1.4B and 2.8B pairs. Context mixing is the main architectural difference within a pair, though training hyperparameters and other parts of the block also differ. Four attention models from other lineages serve as references: Qwen2.5-0.5B, -1.5B and -3B, and SmolLM2-1.7B. All fourteen are base models, run in 16-bit precision through the Hugging Face \texttt{transformers} reference implementations on one 8\,GB consumer GPU, with greedy decoding.

\paragraph{Probe 1a: free recall.} We take 200 SimpleQA questions \cite{wei2024simpleqa}, the first 200 after shuffling the benchmark with seed 42, and pose each to the 2.8B pair after a three-example \texttt{Q:/A:} prompt with a 64-token limit. Two annotators labelled all 400 responses as \emph{correct}, \emph{factual error}, \emph{abstention} or \emph{degenerate}, independently and from a shuffled sheet without model labels; the model key was opened after scoring. Raw agreement was 98.75\% (5 disagreements in 400), Cohen's $\kappa=0.85$.

\paragraph{Probe 1b: context conflict.} The probe covers 111 facts of six types: 56 national capitals, including less familiar ones such as those of Burkina Faso and Kyrgyzstan; 12 chemical symbols; 16 currencies; 15 authors of well-known books; 9 painters of well-known paintings; and 3 facts about planets. Each fact has a true answer and a plausible false one (Canberra and Sydney for the capital of Australia, say). Each fact appears in nine conditions. It is asked once with no context, to record the model's prior answer. In six \emph{conflict} items the false version of the fact is placed first, in the middle or last among unrelated true statements, in a 9-line or a 31-line text. In two \emph{support} items the true version sits in the middle of a 9-line or a 31-line text. This gives 999 items per model. Context items are preceded by two worked examples with invented facts that ask the model to answer only from the text. Decoding is greedy with 12 new tokens. After accent folding and lower-casing, a conflict response is \emph{context-followed} if it contains the planted answer as a whole word and not the true one, and \emph{prior-fallback} in the reverse case. A support response is correct if it contains the true answer.

\paragraph{Controls.} Two confounds could separate the architectures without any difference in how context is weighed. A model might ignore a planted line simply because it cannot read it. Support items on facts the model did not know at baseline test this, since the text is the only route to a correct answer. A model might also resist planted facts because it knows more facts. We therefore repeat each paired test on the facts that both models in the pair answered correctly without context.

\paragraph{Pilot.} An earlier 24-fact version of Probe~1b with nine-line texts was run on the 1.4B and 2.8B pairs. Both 2.8B models followed the planted statement on all 72 context items, while the 1.4B pair separated (70 vs.\ 17 of 72). The ceiling at 2.8B is why the main probe uses longer texts, harder facts and support items (Appendix~\ref{app:pilot}).

\begin{table}[t]
\centering
\footnotesize
\setlength{\tabcolsep}{3.2pt}
\begin{tabular}{@{}lrrrrr@{}}
\toprule
 & Prior & \multicolumn{2}{c}{Repeats false} & Keeps & Reads \\
\cmidrule(lr){3-4}
Model & /111 & 9-line & 31-line & own & unknown \\
\midrule
Pythia-160M & 5 & 55.3\% & 38.4\% & 2.1\% & 25/106 \\
Mamba-130M & 15 & 44.4\% & 36.9\% & 8.7\% & 41/96 \\
\addlinespace
Pythia-410M & 41 & 99.4\% & 94.6\% & 2.1\% & 68/70 \\
Mamba-370M & 54 & 77.5\% & 50.8\% & 20.1\% & 43/57 \\
\addlinespace
Pythia-1B & 27 & 98.8\% & 97.3\% & 2.1\% & 83/84 \\
Mamba-790M & 97 & 91.9\% & 70.9\% & 21.0\% & 14/14 \\
\addlinespace
Pythia-1.4B & 83 & 98.2\% & 96.4\% & 3.3\% & 28/28 \\
Mamba-1.4B & 105 & 26.4\% & 29.4\% & 67.6\% & 4/6 \\
\addlinespace
Pythia-2.8B & 95 & 99.7\% & 95.5\% & 4.2\% & 16/16 \\
Mamba-2.8B & 103 & 93.7\% & 76.3\% & 21.6\% & 7/8 \\
\midrule
Qwen2.5-0.5B & 91 & 99.4\% & 98.8\% & 1.2\% & 20/20 \\
Qwen2.5-1.5B & 108 & 100.0\% & 99.1\% & 0.6\% & 3/3 \\
Qwen2.5-3B & 106 & 100.0\% & 99.4\% & 0.6\% & 5/5 \\
SmolLM2-1.7B & 110 & 100.0\% & 100.0\% & 0.0\% & 1/1 \\
\bottomrule
\end{tabular}
\caption{Probe~1b per model. ``Repeats false'' and ``keeps own'' (answer, 31-line) are shares of the 333 conflict items at each text length. ``Reads unknown'': 31-line support items answered correctly, among facts the model did not know without context. The last four rows are attention models from other lineages.}
\label{tab:s1}
\end{table}

\begin{table*}[t]
\centering
\small
\setlength{\tabcolsep}{5pt}
\begin{tabular}{@{}lrrrcrr@{}}
\toprule
 & \multicolumn{3}{c}{All conflict items ($n=666$)} & & \multicolumn{2}{c}{Facts both knew, 31-line} \\
\cmidrule(lr){2-4}\cmidrule(lr){6-7}
Pair (Pythia / Mamba) & Repeated (P / M) & Discordant (P only / M only) & $p$ & & Repeated (P / M of $n$) & $p$ \\
\midrule
160M / 130M & 312 / 271 & 152 / 111 & 0.014 & & 8 / 14 of 15 & 0.031 \\
410M / 370M & 646 / 427 & 228 / 9 & $5.3\times10^{-56}$ & & 80 / 28 of 87 & $2.3\times10^{-13}$ \\
1B / 790M & 653 / 542 & 113 / 2 & $3.2\times10^{-31}$ & & 73 / 40 of 78 & $2.3\times10^{-10}$ \\
1.4B / 1.4B & 648 / 186 & 462 / 0 & $1.7\times10^{-139}$ & & 235 / 63 of 243 & $3.3\times10^{-52}$ \\
2.8B / 2.8B & 650 / 566 & 90 / 6 & $2.5\times10^{-20}$ & & 264 / 210 of 279 & $9.0\times10^{-13}$ \\
\bottomrule
\end{tabular}
\caption{Probe~1b paired tests (exact McNemar). ``Repeated'': conflict items on which the model gave the planted false answer. The right-hand block repeats the test on 31-line conflict items for facts that both models in the pair answered correctly without context.}
\label{tab:pairs}
\end{table*}

\begin{figure*}[t]
\centering
\includegraphics[width=\textwidth]{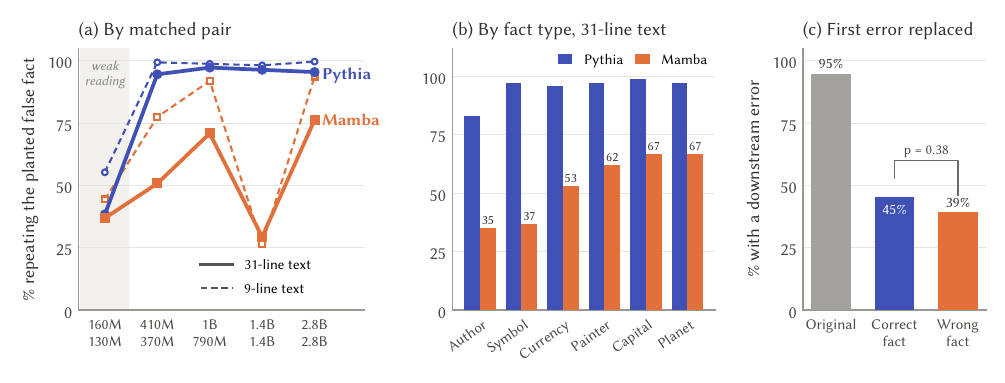}
\caption{(a) Study~1: share of conflict items on which the model repeats the planted false fact, for each Pythia/Mamba pair, with 9-line and 31-line texts. The shaded smallest pair often fails to read the planted line (reading control, Table~\ref{tab:s1}) and is not interpreted. (b) The same outcome by fact type, pooled over the four pairs from 410M to 2.8B, 31-line texts. (c) Study~2, on the 97 items where neither substitution was refused: share of answers with a downstream error. A different wrong fact does as well as the correct one (exact McNemar $p=0.38$).}
\label{fig:results}
\end{figure*}

\begin{figure*}[t]
\centering
\includegraphics[width=\textwidth]{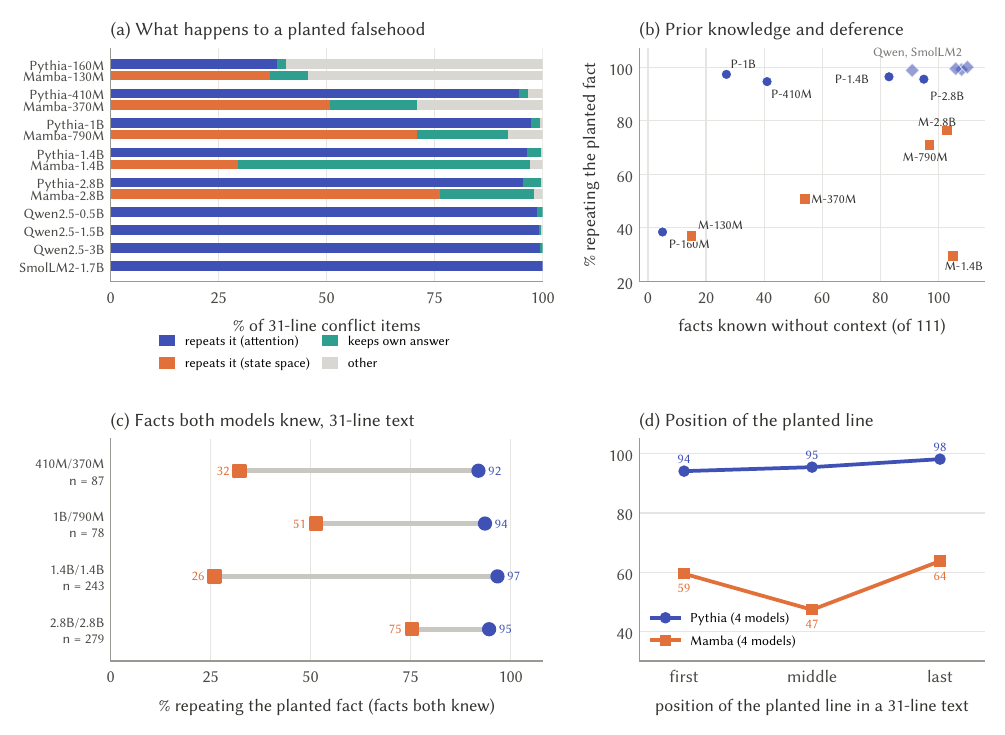}
\caption{Study~1 in more detail, all on 31-line conflict items. (a) Every response sorted into three outcomes: the model repeats the planted fact, gives its own (true) answer, or gives something else. Attention models almost never keep their own answer; the state-space models often do. (b) Facts a model knew without context against how often it repeated the planted falsehood. The attention models stay near the top whatever they know. (c) The paired comparison restricted to facts both models knew; $n$ counts items. (d) Position of the planted line, pooled over the four pairs from 410M to 2.8B (444 items per point).}
\label{fig:s1detail}
\end{figure*}

\subsection{Results}

\paragraph{Free recall.} Each 2.8B model answered 3 of 200 questions correctly, and both produced a factual error on 182 of the same questions. Pythia abstained or degenerated on 10 items and Mamba on 2, which accounts for most of the difference in factual-error counts (exact McNemar on the factual-error indicator, $p=0.096$). With data held fixed, removing attention does not change recall failure on long-tail questions. Both models are near floor, so the probe cannot rule out a smaller role for attention in recall.

\paragraph{Context conflict.} Every attention model from 410M upwards repeated the planted false fact on at least 94.6\% of 31-line conflict items, and the four reference models on at least 98.8\% (Table~\ref{tab:s1}). The Mamba models did so far less often: 50.8\% at 370M, 70.9\% at 790M, 29.4\% at 1.4B and 76.3\% at 2.8B. When Mamba did not follow the text, it mostly gave its own answer, on 20--68\% of 31-line conflict items against at most 4.2\% for Pythia (Figure~\ref{fig:s1detail}a). Mamba-370M is the exception: on 29\% of items it gave neither answer. In each of the four pairs from 410M to 2.8B the difference is large and runs in one direction (Table~\ref{tab:pairs}); at 2.8B, 90 items were repeated only by Pythia and 6 only by Mamba ($p=2.5\times10^{-20}$). Across all fourteen models the 31-line rate is higher in the attention group (mean 91.1\%, nine models) than in the state-space group (mean 52.9\%, five models; Mann--Whitney $p=0.007$).

\paragraph{Controls.} The gap is not a reading failure. On facts it did not know without context, Mamba-790M repeated the planted true fact on 14 of 14 support items and Mamba-2.8B on 7 of 8 (8 of 8 with 9-line texts), so these models retrieve a planted line when it is the only source of the answer. Nor is it explained by Mamba knowing more facts: restricted to facts both models knew, every pair from 410M to 2.8B still separates in the same direction ($p\le2.3\times10^{-10}$). On those known facts the attention models still repeated the falsehood on 93.5--100\% of 31-line items. Figure~\ref{fig:s1detail}b makes the same point across all fourteen models: how many facts a model knows says little about whether an attention model will give way to the text. The smallest pair behaves differently. Pythia-160M and Mamba-130M answered the support items correctly on only 25 of 106 and 41 of 96 unknown facts; they often cannot read the planted line, and we do not interpret that pair.

\subsection{Error Analysis}

\paragraph{Text length and position.} Pythia's rate barely moves with text length: from 410M upwards it is at least 98.2\% with 9-line texts and at least 94.6\% with 31-line texts. Mamba's rate falls as the text grows, at 2.8B from 93.7\% to 76.3\% and at 790M from 91.9\% to 70.9\%. Position matters as well (P1), but mainly for Mamba. With 31-line texts, Mamba-2.8B followed the planted line on 79.3\% of items when it came first, 63.1\% in the middle and 86.5\% last; Pythia-2.8B on 94.6\%, 93.7\% and 98.2\%. Pooled over the four pairs from 410M to 2.8B, the Mamba models followed it on 59.5\%, 47.3\% and 63.7\% of items, and the Pythia models on 94.1\%, 95.5\% and 98.2\% (Figure~\ref{fig:s1detail}d). The state-space models thus show the middle dip that \newcite{liu2024} report for attention models over much longer contexts. At this text length the attention models show no dip, only a slight rise when the planted line comes last.

\paragraph{Fact type.} Pooled over the four interpretable pairs, Pythia repeated the planted fact on 83--99\% of items of every type (Figure~\ref{fig:results}b). Mamba's rate depends on the type: 35\% for authors of books and 37\% for chemical symbols, against 67\% for capitals and planets. The types where Mamba most often keeps its own answer are those where the answer is a name or symbol bound tightly to its subject in text (\emph{Pride and Prejudice} and Austen, gold and Au). Authors are also where Pythia is least compliant (83\%). Strong associations thus raise resistance to a planted falsehood in both architectures, far more so in the state-space model.

\subsection{Interpretation and Limits}
\label{sec:s1-limits}

The results meet P5. Replacing attention with a state-space mixer, with data, tokenizer and scale matched, left free-recall failure unchanged and reduced the adoption of misleading context at every interpretable scale. The controls rule out the simplest alternatives: the Mamba models can read the planted line, and the gap holds on facts both models knew. In these models, attention-based mixing goes with near-unconditional deference to the context, right or wrong. The state-space models weigh the text against what they already know, more so as the text grows longer. Read through the data-as-amplifier claim of Section~\ref{sec:framework}, the same defect, a false line in the input, became an output error at very different rates depending on the mixer it passed through.

Four limits apply. The architectures differ in more than context mixing, so the effect belongs to the architecture family rather than to attention in isolation. State-space models are known to be weaker at in-context recall \cite{arora2024zoology,jelassi2024repeat}; our reading control shows the planted line was read, but cannot exclude that it was read less reliably when it competed with a stored fact. The reading control at 2.8B rests on 8 unknown facts, because Mamba-2.8B knew 103 of 111. The facts, texts and fillers are synthetic and in English, and scoring is automatic string matching, which counts a response containing both answers as neither followed nor fallback.

\section{Study 2: Repairing the First Error, With Controls}
\label{sec:study2}

\subsection{Setup}

\paragraph{Items and models.} We drew a simple random sample of 200 SimpleQA questions with a fixed seed (20260819). Three instruction-tuned models served through the Groq API answered every question at temperature 0 with the system prompt ``Answer the question directly and completely'': GPT-OSS-120B, Qwen3.6-27B (which produces a reasoning trace before answering) and ALLaM-2-7B. Of the 600 responses, 128 stated a wrong fact and followed it with at least one claim building on it. We removed 17 in which the model hedged or refused instead of committing to a single wrong fact (15 from ALLaM-2-7B, 2 from GPT-OSS-120B), leaving 111 items: 72 from ALLaM-2-7B, 21 from Qwen3.6-27B and 18 from GPT-OSS-120B. For each item we recorded the character offset of the first divergence from a correct answer and wrote two replacements for the divergent span: the SimpleQA gold answer, and a plausible wrong alternative that differs from the model's own error.

\paragraph{Conditions.} \textbf{A (baseline)} is the original response. In \textbf{B (correct substitution)} the response up to the divergence, followed by the gold answer, is supplied as a partial assistant turn and the model continues from it. \textbf{D (wrong substitution)} is the same with the alternative wrong fact. \textbf{C (isolated)} asks the question again with no prior answer, using the fixed template ``In one clear sentence, what is the correct answer to this: [question]''.

\paragraph{Scoring.} For A, B and D, two annotators counted the claims after the divergence point that the reference contradicts or does not support, and flagged abstentions; for C they scored whether the answer was correct. Both scored all 444 rows independently and blind to condition. Agreement on the presence of any downstream error was $\kappa=0.94$ ($n=333$; exact agreement on the count 91.6\%), and $\kappa=1.0$ on isolated correctness ($n=111$) and on abstention ($n=222$). The main outcome is whether a response contains at least one downstream error. A refusal under B or D contains no claims, so we treat it as missing rather than error-free.

\paragraph{Hypotheses.} The hypotheses, threshold and tests were fixed before scoring. H1: B reduces the downstream-error rate relative to A by at least 20 percentage points. H2: on items where B removes all downstream errors, the model answers C correctly. H3: B yields fewer downstream errors than D. H1 tests P3; H2 and H3 test P3b and P3a.

\paragraph{Deviations from the plan.} There are six. (1) The original pool of self-written questions produced 0--1 qualifying errors and was replaced with SimpleQA. (2) Qwen3.6-27B's token limit was raised from 1024 to 2048 because its reasoning traces were cut off; 47 of its 200 responses still ended inside the reasoning trace and were excluded. (3) The isolated query was reworded with the fixed template above, because at temperature 0 the baseline wording reproduces the baseline answer. (4) The planned test for H3 was a two-proportion $z$-test; B and D are measured on the same items, so we report the exact McNemar test as primary and the planned test beside it. (5) A first analysis indexed items by question only. Seventeen questions produced qualifying errors from two models, so one model's rows overwrote the other's and 90 items remained. The corrected analysis indexes by question and model and uses all 111; both scripts are released. (6) Using one annotator's scores for the tests and the other's for reliability was not specified before scoring; the second annotator's scores lead to the same conclusions (Appendix~\ref{app:s2second}).

\begin{table*}[t]
\centering
\small
\renewcommand{\arraystretch}{1.1}
\begin{tabular}{@{}lL{5.2cm}lL{6.3cm}@{}}
\toprule
 & Comparison & Result & Test \\
\midrule
H1 & A vs.\ B, share with an error ($n=106$) & 92.5\% $\to$ 48.1\% & exact McNemar, 50 vs.\ 3 discordant, $p=5.5\times10^{-12}$ \\
 & A vs.\ B, failing claims per answer & 1.70 $\to$ 0.75 & Wilcoxon signed-rank, $p=1.8\times10^{-11}$ \\
\addlinespace
H2 & C correct where B succeeded ($n=50$) & 1/50 (2.0\%) & Wilson 95\% CI [0.4\%, 10.5\%] \\
\addlinespace
H3 & B vs.\ D, share error-free ($n=97$) & 54.6\% vs.\ 60.8\% & exact McNemar, 13 vs.\ 19 discordant, $p=0.38$; planned $z$-test, $z=-0.87$, $p=0.38$ \\
 & B vs.\ D, failing claims per answer & 0.66 vs.\ 0.76 & Wilcoxon signed-rank, $p=0.91$ \\
\midrule
 & C correct, all items ($n=111$) & 2/111 (1.8\%) & -- \\
 & Self-correction within A ($n=111$) & 0/111 & -- \\
 & Refusals under B / D ($n=111$) & 5 / 12 & exact McNemar, 2 vs.\ 9 discordant, $p=0.065$ \\
\bottomrule
\end{tabular}
\caption{Study~2 results (primary annotator). A downstream error is at least one failing claim after the divergence point.}
\label{tab:s2}
\end{table*}

\subsection{Results}

\paragraph{H1 holds.} With the correct fact inserted, the share of answers containing a downstream error fell from 92.5\% to 48.1\%, a 44.3-point drop against a planned threshold of 20 points (Table~\ref{tab:s2}; Benjamini--Hochberg $q=1.1\times10^{-11}$). The mean number of failing claims per answer fell from 1.70 to 0.75. The direction is the same for each model: ALLaM-2-7B from 94.2\% to 50.7\% ($n=69$, $p=1.5\times10^{-8}$), Qwen3.6-27B from 100\% to 42.1\% ($n=19$, $p=0.001$), GPT-OSS-120B from 77.8\% to 44.4\% ($n=18$, $p=0.11$). No baseline answer corrected itself.

\paragraph{H2 and H3 do not hold.} Asked directly, the models gave the correct answer on 2 of 111 items, both from ALLaM-2-7B, and on 1 of the 50 items where the correct fact removed every downstream error. Inserting a different wrong fact reduced downstream errors about as much as inserting the correct one. On the binary outcome the wrong fact did slightly better (60.8\% vs.\ 54.6\% error-free); on the claim count the correct fact did slightly better (0.66 vs.\ 0.76 failing claims). Neither difference is significant. Refusals were more frequent after a wrong substitution (12) than after a correct one (5), short of significance ($p=0.065$).

\begin{figure*}[t]
\centering
\includegraphics[width=\textwidth]{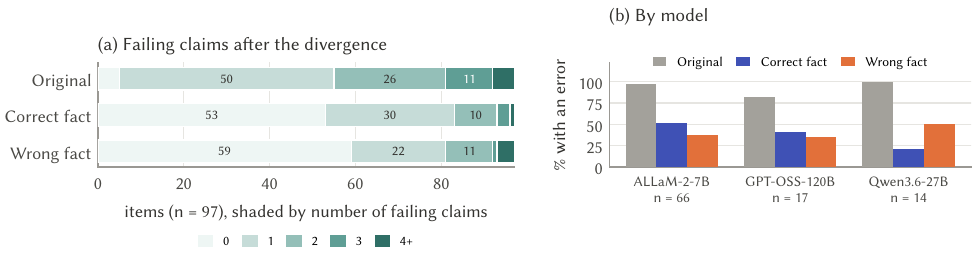}
\caption{Study~2 on the 97 items where neither substitution was refused. (a) Number of failing claims after the divergence point. (b) Share of answers with at least one downstream error, by model.}
\label{fig:s2detail}
\end{figure*}

\paragraph{Where the errors go.} Figure~\ref{fig:s2detail}a shows the whole distribution rather than the binary outcome. Both substitutions mostly turn one- and two-claim cascades into clean continuations, and they shrink but do not remove the tail of answers with three or more failing claims (16 in the original answers, 4 after the correct fact and 5 after the wrong one). By model (Figure~\ref{fig:s2detail}b), the correct fact did clearly better than the wrong one only for Qwen3.6-27B (3 vs.\ 7 of 14 answers with an error), too few items to test on their own.

\paragraph{Length.} Fewer errors could simply mean shorter continuations. Measured from the divergence point, the median continuation is 29 words in A, 28 in B and 29 in D.

\subsection{Interpretation}

H1 shows that what follows depends on what was committed: change the committed fact and about half of the downstream errors go away. This is the behavioural signature of commitment under exposure bias, and it holds in a reasoning model as well as in two conventional instruction-tuned ones. H2 and H3 rule out the stronger explanation on these items. The models did not know the facts, and whether the inserted fact was true made no detectable difference. \newcite{zhang2024snowball} built their items from facts the models know, and knowledge-based commitment may operate there. Our items are long-tail SimpleQA questions, where the objective and sparse data already predict failure \cite{kandpal2023}, so the knowledge route is closed. Two explanations remain and the design does not separate them: any substitution may break a chain of elaboration that the original fact supported, or some downstream claims in the baseline may be independent fabrications that a new prefix happens to produce less often. Under the T1 attribution rule, which requires both a successful repair and correct isolated knowledge, 1 of 111 errors is attributed to commitment.

\section{Discussion}
\label{sec:discussion}

\paragraph{What survives of the structural account.} Two of its three parts were tested, and each came out differently. The attention part holds in a specific form: at matched data and scale, attention models adopted a planted falsehood almost regardless of what they knew, and state-space models kept their own answer far more often while still reading the text. This is what ``associative retrieval without a validity check'' predicts, and it is now a measured property of two architecture families rather than an argument. The commitment part holds behaviourally but not mechanistically: the first committed fact steers what follows, yet not because a known truth was suppressed. The objective part was not tested directly, but two results point to it. Free-recall failure was the same with and without attention, and on the long-tail questions of Study~2 the models almost never knew the answer. A separate pre-registered evaluation of five open-weight models \cite{sadi2026wrong} fits the same picture: questions from niche domains were answered wrongly more often than general-domain ones (65.5\% vs.\ 43.3\%, odds ratio 2.49).

\paragraph{Where data comes in.} The original account said that data amplifies hallucination without originating it. Our results support a narrower version and do not support the stronger one. In Study~1 the defect was held fixed and the architecture varied, and the defect turned into an error at very different rates: at least 93\% of 31-line conflict items on known facts for every attention model from 410M upwards, 29--76\% for the Mamba models. In Study~2 the first error came from sparse coverage of the fact, and the decoding process then elaborated on it. The components thus decide whether a defect reaches the output and how far an error spreads; data decides how often there is a defect to pass on. None of this shows that architecture explains more of the variance in hallucination than data does, and we do not claim it.

\paragraph{The modern training stack.} Deployed models are instruction-tuned and preference-optimised \cite{ouyang2022,rafailov2023}, and the strict claim that the objective contains no truth term applies to pretraining only. Two observations limit how much this changes. First, preference training rewards predicted approval, not truth, and preference-trained assistants defer to the user's stated view \cite{sharma2024sycophancy}, which is the conversational analogue of the context deference measured in Study~1. Second, the three models in Study~2 are all instruction-tuned, and the commitment effect appears in each of them, reasoning trace included. Study~1, by contrast, uses base models of at most 3B parameters, and whether instruction tuning strengthens or weakens context deference in matched architectures is open. On scale, deference in the attention models saturates early: from 410M upwards every Pythia model was at 94.6\% or above, and the Qwen and SmolLM2 models at 98.8\% or above. Mamba's resistance did not grow monotonically with size (50.8\%, 70.9\%, 29.4\%, 76.3\% from 370M to 2.8B), so in this range scale does not remove the difference.

\paragraph{Retrieval and calibration.} For retrieval-augmented systems \cite{lewis2020}, the attention models, the architecture behind most deployed systems, repeated a planted falsehood on at least 93\% of 31-line conflict items even for facts they knew. This matches reports that a few injected passages can steer retrieval-augmented answers \cite{zou2025poisonedrag} and that models side with external evidence over their prior across many domains \cite{wu2024clasheval}. Retrieval changes what the model reads, not how it reads; filtering what enters the context therefore matters more than the model's own knowledge. State-space models resist more, but that resistance also costs them when the context is the better source, and hybrid attention/state-space models are an obvious next case. On calibration, the companion evaluation found that 77.4\% of incorrect answers carried no hedge \cite{sadi2026wrong}. Probing work shows that a correctness signal often exists inside the model when the output does not express it \cite{azaria2023,orgad2025,farquhar2024}. Our interventions are complementary: they work from the output alone and can route an error to a component, which tells internal-state methods where to look.

\paragraph{Controls decide the label.} Both studies illustrate the gap between performance and mechanism that \newcite{callegari2026} describe. In Study~2 a large, significant behavioural effect invited a mechanistic reading that a cheap control showed to be wrong. In Study~1 the pilot was inconclusive: with familiar facts and short texts both 2.8B models sat at ceiling. The attribution became supportable only once the probe was hard enough to avoid the ceiling and included controls for reading and for prior knowledge. We do not conclude that component attribution from outputs should be given up. We conclude that it should be conditional on specific controls: matched comparisons at more than one scale for architectural claims, counterfactual inputs that differ only in truth value for claims about knowledge, and a check that the model can use the input at all.

\paragraph{A component axis for error analysis.} Error analysis in NLG is still uncommon \cite{vanmiltenburg2023error}. Recording, for each error, both its output type \cite{alansari2025} and the component an intervention attributes it to would make such analyses more useful for deciding what to fix. Two extrinsic errors that look the same may call for different remedies: one disappears when a misleading passage is removed, the other persists under a minimal prompt. As with the effect-size approach of \newcite{dayanik2022}, the second label should be assigned only after confounds have been controlled.

\section{Limitations}

Study~1 compares architecture families that differ in more than context mixing, uses synthetic English texts and automatic scoring, and has a reading control at 2.8B that rests on 8 facts. All Study~1 models are base models of at most 3B parameters. Study~2 uses long-tail questions on which the models almost never know the answer; this produced enough errors to analyse, but it is the setting least favourable to knowledge-based commitment, and an item set restricted to questions each model answers correctly in isolation is the direct follow-up. ALLaM-2-7B supplies 72 of 111 items, so per-model estimates for the other two models are imprecise. Of the three interventions, T1 and T2 were run; the frequency test T3, a classifier baseline for attribution labels, and a test of the data claim with pretraining data actually varied are left for future work. Some of the annotators in Studies~1 and~2 are also authors; blind scoring and high agreement reduce the risk of expectation effects but do not remove it.

\section{Conclusion}

Hallucination is usually classified by the type of output and explained by appeal to the model's structure. We tested that structural explanation with interventions that need only sampling access. Across five Pythia/Mamba pairs matched on data and scale, removing attention left free-recall failure unchanged but reduced the adoption of misleading context at every interpretable scale, with reading ability and prior knowledge controlled. In three instruction-tuned models, replacing the first error in a long-tail answer halved the errors that followed whether or not the replacement was true, and the models rarely knew the correct answer. The structural account holds in a narrower form than it is usually stated. Context mixing decides whether a defect in the input becomes an output error, commitment decides how far an error spreads, and on long-tail questions the error itself comes from missing knowledge. Data supplies the defects; the components decide what becomes of them. Component labels for hallucination should be assigned only after controls of the kind used here.

\section*{Ethical Considerations}
The work aims to make confident model errors easier to diagnose. It uses public benchmarks and publicly available models and involves no personal data. Probe~1b confirms that models readily adopt false statements placed in their context. This is already well documented, and nothing is released for producing such inputs beyond the probe items, which concern well-known public facts.

\bibliography{references}

@inproceedings{maynez2020,
  author    = {Maynez, Joshua and Narayan, Shashi and Bohnet, Bernd and McDonald, Ryan},
  title     = {On Faithfulness and Factuality in Abstractive Summarization},
  booktitle = {Proceedings of the 58th Annual Meeting of the Association for Computational Linguistics},
  pages     = {1906--1919},
  year      = {2020}
}

@article{ji2023,
  author  = {Ji, Ziwei and Lee, Nayeon and Frieske, Rita and Yu, Tiezheng and Su, Dan and Xu, Yan and Ishii, Etsuko and Bang, Ye Jin and Madotto, Andrea and Fung, Pascale},
  title   = {Survey of Hallucination in Natural Language Generation},
  journal = {ACM Computing Surveys},
  volume  = {55},
  number  = {12},
  pages   = {1--38},
  year    = {2023}
}

@article{huang2023,
  author  = {Huang, Lei and Yu, Weijiang and Ma, Weitao and Zhong, Weihong and Feng, Zhangyin and Wang, Haotian and Chen, Qianglong and Peng, Weihua and Feng, Xiaocheng and Qin, Bing and Liu, Ting},
  title   = {A Survey on Hallucination in Large Language Models: Principles, Taxonomy, Challenges, and Open Questions},
  journal = {ACM Transactions on Information Systems},
  volume  = {43},
  number  = {2},
  pages   = {1--55},
  year    = {2025}
}

@article{alansari2025,
  author  = {Alansari, Aisha and Luqman, Hamzah},
  title   = {Large Language Models Hallucination: A Comprehensive Survey},
  journal = {Computer Science Review},
  volume  = {61},
  pages   = {100970},
  year    = {2026}
}

@inproceedings{dziri2022,
  author    = {Dziri, Nouha and Milton, Sivan and Yu, Mo and Zaiane, Osmar and Reddy, Siva},
  title     = {On the Origin of Hallucinations in Conversational Models: Is it the Datasets or the Models?},
  booktitle = {Proceedings of the 2022 Conference of the North American Chapter of the Association for Computational Linguistics: Human Language Technologies},
  pages     = {5271--5285},
  year      = {2022}
}

@inproceedings{mckenna2023,
  author    = {McKenna, Nick and Li, Tianyi and Cheng, Liang and Hosseini, Mohammad Javad and Johnson, Mark and Steedman, Mark},
  title     = {Sources of Hallucination by Large Language Models on Inference Tasks},
  booktitle = {Findings of the Association for Computational Linguistics: EMNLP 2023},
  year      = {2023}
}

@inproceedings{kandpal2023,
  author    = {Kandpal, Nikhil and Deng, Haikang and Roberts, Adam and Wallace, Eric and Raffel, Colin},
  title     = {Large Language Models Struggle to Learn Long-Tail Knowledge},
  booktitle = {Proceedings of the 40th International Conference on Machine Learning},
  series    = {PMLR},
  volume    = {202},
  year      = {2023}
}

@inproceedings{mallen2023,
  author    = {Mallen, Alex and Asai, Akari and Zhong, Victor and Das, Rajarshi and Khashabi, Daniel and Hajishirzi, Hannaneh},
  title     = {When Not to Trust Language Models: Investigating Effectiveness of Parametric and Non-Parametric Memories},
  booktitle = {Proceedings of the 61st Annual Meeting of the Association for Computational Linguistics},
  year      = {2023}
}

@article{kalai2025,
  author  = {Kalai, Adam Tauman and Nachum, Ofir and Vempala, Santosh S. and Zhang, Edwin},
  title   = {Why Language Models Hallucinate},
  journal = {arXiv preprint arXiv:2509.04664},
  year    = {2025}
}

@inproceedings{geva2021,
  author    = {Geva, Mor and Schuster, Roei and Berant, Jonathan and Levy, Omer},
  title     = {Transformer Feed-Forward Layers Are Key-Value Memories},
  booktitle = {Proceedings of the 2021 Conference on Empirical Methods in Natural Language Processing},
  pages     = {5484--5495},
  year      = {2021}
}

@inproceedings{geva2023,
  author    = {Geva, Mor and Bastings, Jasmijn and Filippova, Katja and Globerson, Amir},
  title     = {Dissecting Recall of Factual Associations in Auto-Regressive Language Models},
  booktitle = {Proceedings of the 2023 Conference on Empirical Methods in Natural Language Processing},
  year      = {2023}
}

@inproceedings{meng2022,
  author    = {Meng, Kevin and Bau, David and Andonian, Alex and Belinkov, Yonatan},
  title     = {Locating and Editing Factual Associations in {GPT}},
  booktitle = {Advances in Neural Information Processing Systems},
  volume    = {35},
  pages     = {17359--17372},
  year      = {2022}
}

@inproceedings{li2023iti,
  author    = {Li, Kenneth and Patel, Oam and Vi{\'e}gas, Fernanda and Pfister, Hanspeter and Wattenberg, Martin},
  title     = {Inference-Time Intervention: Eliciting Truthful Answers from a Language Model},
  booktitle = {Advances in Neural Information Processing Systems},
  volume    = {36},
  year      = {2023}
}

@article{olsson2022,
  author  = {Olsson, Catherine and Elhage, Nelson and Nanda, Neel and Joseph, Nicholas and DasSarma, Nova and Henighan, Tom and Mann, Ben and Askell, Amanda and Bai, Yuntao and Chen, Anna and Conerly, Tom and Drain, Dawn and Ganguli, Deep and Hatfield-Dodds, Zac and Hernandez, Danny and Johnston, Scott and Jones, Andy and Kernion, Jackson and Lovitt, Liane and Ndousse, Kamal and Amodei, Dario and Brown, Tom and Clark, Jack and Kaplan, Jared and McCandlish, Sam and Olah, Chris},
  title   = {In-Context Learning and Induction Heads},
  journal = {Transformer Circuits Thread},
  year    = {2022},
  url     = {https://transformer-circuits.pub/2022/in-context-learning-and-induction-heads/index.html}
}

@inproceedings{azaria2023,
  author    = {Azaria, Amos and Mitchell, Tom},
  title     = {The Internal State of an {LLM} Knows When It's Lying},
  booktitle = {Findings of the Association for Computational Linguistics: EMNLP 2023},
  pages     = {967--976},
  year      = {2023}
}

@inproceedings{orgad2025,
  author    = {Orgad, Hadas and Toker, Michael and Gekhman, Zorik and Reichart, Roi and Szpektor, Idan and Kotek, Hadas and Belinkov, Yonatan},
  title     = {{LLMs} Know More Than They Show: On the Intrinsic Representation of {LLM} Hallucinations},
  booktitle = {International Conference on Learning Representations},
  year      = {2025}
}

@article{farquhar2024,
  author  = {Farquhar, Sebastian and Kossen, Jannik and Kuhn, Lorenz and Gal, Yarin},
  title   = {Detecting Hallucinations in Large Language Models Using Semantic Entropy},
  journal = {Nature},
  volume  = {630},
  number  = {8017},
  pages   = {625--630},
  year    = {2024}
}

@inproceedings{ranzato2016,
  author    = {Ranzato, Marc'Aurelio and Chopra, Sumit and Auli, Michael and Zaremba, Wojciech},
  title     = {Sequence Level Training with Recurrent Neural Networks},
  booktitle = {International Conference on Learning Representations},
  year      = {2016}
}

@inproceedings{bengio2015,
  author    = {Bengio, Samy and Vinyals, Oriol and Jaitly, Navdeep and Shazeer, Noam},
  title     = {Scheduled Sampling for Sequence Prediction with Recurrent Neural Networks},
  booktitle = {Advances in Neural Information Processing Systems},
  volume    = {28},
  year      = {2015}
}

@inproceedings{wang2020exposure,
  author    = {Wang, Chaojun and Sennrich, Rico},
  title     = {On Exposure Bias, Hallucination and Domain Shift in Neural Machine Translation},
  booktitle = {Proceedings of the 58th Annual Meeting of the Association for Computational Linguistics},
  pages     = {3544--3552},
  year      = {2020}
}

@inproceedings{he2021exposure,
  author    = {He, Tianxing and Zhang, Jingzhao and Zhou, Zhiming and Glass, James},
  title     = {Exposure Bias versus Self-Recovery: Are Distortions Really Incremental for Autoregressive Text Generation?},
  booktitle = {Proceedings of the 2021 Conference on Empirical Methods in Natural Language Processing},
  year      = {2021}
}

@inproceedings{zhang2024snowball,
  author    = {Zhang, Muru and Press, Ofir and Merrill, William and Liu, Alisa and Smith, Noah A.},
  title     = {How Language Model Hallucinations Can Snowball},
  booktitle = {Proceedings of the 41st International Conference on Machine Learning},
  pages     = {59670--59684},
  year      = {2024}
}

@inproceedings{longpre2021,
  author    = {Longpre, Shayne and Perisetla, Kartik and Chen, Anthony and Ramesh, Nikhil and DuBois, Chris and Singh, Sameer},
  title     = {Entity-Based Knowledge Conflicts in Question Answering},
  booktitle = {Proceedings of the 2021 Conference on Empirical Methods in Natural Language Processing},
  year      = {2021}
}

@inproceedings{xie2024,
  author    = {Xie, Jian and Zhang, Kai and Chen, Jiangjie and Lou, Renze and Su, Yu},
  title     = {Adaptive Chameleon or Stubborn Sloth: Revealing the Behavior of Large Language Models in Knowledge Conflicts},
  booktitle = {International Conference on Learning Representations},
  year      = {2024}
}

@inproceedings{shi2023distracted,
  author    = {Shi, Freda and Chen, Xinyun and Misra, Kanishka and Scales, Nathan and Dohan, David and Chi, Ed H. and Sch{\"a}rli, Nathanael and Zhou, Denny},
  title     = {Large Language Models Can Be Easily Distracted by Irrelevant Context},
  booktitle = {Proceedings of the 40th International Conference on Machine Learning},
  year      = {2023}
}

@article{liu2024,
  author  = {Liu, Nelson F. and Lin, Kevin and Hewitt, John and Paranjape, Ashwin and Bevilacqua, Michele and Petroni, Fabio and Liang, Percy},
  title   = {Lost in the Middle: How Language Models Use Long Contexts},
  journal = {Transactions of the Association for Computational Linguistics},
  volume  = {12},
  pages   = {157--173},
  year    = {2024}
}

@inproceedings{gu2023mamba,
  author    = {Gu, Albert and Dao, Tri},
  title     = {Mamba: Linear-Time Sequence Modeling with Selective State Spaces},
  booktitle = {First Conference on Language Modeling},
  year      = {2024}
}

@inproceedings{biderman2023pythia,
  author    = {Biderman, Stella and Schoelkopf, Hailey and Anthony, Quentin and Bradley, Herbie and O'Brien, Kyle and Hallahan, Eric and Khan, Mohammad Aflah and Purohit, Shivanshu and Prashanth, USVSN Sai and Raff, Edward and Skowron, Aviya and Sutawika, Lintang and van der Wal, Oskar},
  title     = {Pythia: A Suite for Analyzing Large Language Models Across Training and Scaling},
  booktitle = {Proceedings of the 40th International Conference on Machine Learning},
  year      = {2023}
}

@article{gao2020pile,
  author  = {Gao, Leo and Biderman, Stella and Black, Sid and Golding, Laurence and Hoppe, Travis and Foster, Charles and Phang, Jason and He, Horace and Thite, Anish and Nabeshima, Noa and Presser, Shawn and Leahy, Connor},
  title   = {The {Pile}: An 800{GB} Dataset of Diverse Text for Language Modeling},
  journal = {arXiv preprint arXiv:2101.00027},
  year    = {2020}
}

@inproceedings{arora2024zoology,
  author    = {Arora, Simran and Eyuboglu, Sabri and Timalsina, Aman and Johnson, Isys and Poli, Michael and Zou, James and Rudra, Atri and R{\'e}, Christopher},
  title     = {Zoology: Measuring and Improving Recall in Efficient Language Models},
  booktitle = {International Conference on Learning Representations},
  year      = {2024}
}

@inproceedings{jelassi2024repeat,
  author    = {Jelassi, Samy and Brandfonbrener, David and Kakade, Sham M. and Malach, Eran},
  title     = {Repeat After Me: Transformers are Better than State Space Models at Copying},
  booktitle = {Proceedings of the 41st International Conference on Machine Learning},
  year      = {2024}
}

@misc{wei2024simpleqa,
  author       = {Wei, Jason and Nguyen, Karina and Chung, Hyung Won and Jiao, Yunxin Joy and Papay, Spencer and Glaese, Amelia and Schulman, John and Fedus, William},
  title        = {Measuring Short-Form Factuality in Large Language Models},
  howpublished = {arXiv preprint arXiv:2411.04368},
  year         = {2024}
}

@inproceedings{lin2022,
  author    = {Lin, Stephanie and Hilton, Jacob and Evans, Owain},
  title     = {{TruthfulQA}: Measuring How Models Mimic Human Falsehoods},
  booktitle = {Proceedings of the 60th Annual Meeting of the Association for Computational Linguistics},
  pages     = {3214--3252},
  year      = {2022}
}

@inproceedings{vaswani2017,
  author    = {Vaswani, Ashish and Shazeer, Noam and Parmar, Niki and Uszkoreit, Jakob and Jones, Llion and Gomez, Aidan N. and Kaiser, {\L}ukasz and Polosukhin, Illia},
  title     = {Attention Is All You Need},
  booktitle = {Advances in Neural Information Processing Systems},
  volume    = {30},
  year      = {2017}
}

@inproceedings{lewis2020,
  author    = {Lewis, Patrick and Perez, Ethan and Piktus, Aleksandra and Petroni, Fabio and Karpukhin, Vladimir and Goyal, Naman and K{\"u}ttler, Heinrich and Lewis, Mike and Yih, Wen-tau and Rockt{\"a}schel, Tim and Riedel, Sebastian and Kiela, Douwe},
  title     = {Retrieval-Augmented Generation for Knowledge-Intensive {NLP} Tasks},
  booktitle = {Advances in Neural Information Processing Systems},
  volume    = {33},
  pages     = {9459--9474},
  year      = {2020}
}

@article{callegari2026,
  author  = {Callegari, Elena and Simonsen, Annika},
  title   = {Rethinking Linguistic Tests for Large Language Models: Airplanes Aren't Birds},
  journal = {Northern European Journal of Language Technology},
  volume  = {12},
  number  = {1},
  year    = {2026}
}

@article{pikuliak2024,
  author  = {Pikuliak, Mat{\'u}{\v{s}}},
  title   = {On Using Self-Report Studies to Analyze Language Models},
  journal = {Northern European Journal of Language Technology},
  volume  = {10},
  number  = {1},
  year    = {2024}
}

@article{leino2025,
  author  = {Leino, Julius and Karlgren, Jussi},
  title   = {Controlling Language and Style of Multi-lingual Generative Language Models with Control Vectors},
  journal = {Northern European Journal of Language Technology},
  volume  = {11},
  number  = {1},
  year    = {2025}
}

@article{vanmiltenburg2023error,
  author  = {van Miltenburg, Emiel and Clinciu, Miruna and Du{\v{s}}ek, Ond{\v{r}}ej and Gkatzia, Dimitra and Inglis, Stephanie and Lepp{\"a}nen, Leo and Mahamood, Saad and Schoch, Stephanie and Thomson, Craig and Wen, Luou},
  title   = {Barriers and Enabling Factors for Error Analysis in {NLG} Research},
  journal = {Northern European Journal of Language Technology},
  volume  = {9},
  number  = {1},
  year    = {2023}
}

@article{vanmiltenburg2023registered,
  author  = {van Miltenburg, Emiel},
  title   = {Resource Papers as Registered Reports: A Proposal},
  journal = {Northern European Journal of Language Technology},
  volume  = {9},
  number  = {1},
  year    = {2023}
}

@article{dayanik2022,
  author  = {Dayanik, Erenay and Vu, Ngoc Thang and Pad{\'o}, Sebastian},
  title   = {Bias Identification and Attribution in {NLP} Models With Regression and Effect Sizes},
  journal = {Northern European Journal of Language Technology},
  volume  = {8},
  number  = {1},
  year    = {2022}
}

@misc{sadi2026wrong,
  author       = {Sadi, Md. Rejaul Korim and Tasin, Toufiqur Rahman and Naeem, Golam Mostofa and Chowdhury, Dewan Fahad and Chowdhury, Hatim Al Amin},
  title        = {Wrong with Confidence: How Large Language Models Hallucinate, Manipulate, and Discriminate},
  howpublished = {Zenodo preprint, doi:10.5281/zenodo.20677241},
  year         = {2026}
}

@inproceedings{sharma2024sycophancy,
  title     = {Towards Understanding Sycophancy in Language Models},
  author    = {Sharma, Mrinank and Tong, Meg and Korbak, Tomasz and Duvenaud, David and Askell, Amanda and Bowman, Samuel R. and Durmus, Esin and Hatfield-Dodds, Zac and Johnston, Scott R. and Kravec, Shauna M. and Maxwell, Timothy and McCandlish, Sam and Ndousse, Kamal and Rausch, Oliver and Schiefer, Nicholas and Yan, Da and Zhang, Miranda and Perez, Ethan},
  booktitle = {The Twelfth International Conference on Learning Representations},
  year      = {2024}
}

@inproceedings{xu2024conflicts,
  title     = {Knowledge Conflicts for {LLM}s: A Survey},
  author    = {Xu, Rongwu and Qi, Zehan and Guo, Zhijiang and Wang, Cunxiang and Wang, Hongru and Zhang, Yue and Xu, Wei},
  booktitle = {Proceedings of the 2024 Conference on Empirical Methods in Natural Language Processing},
  pages     = {8541--8565},
  year      = {2024},
  publisher = {Association for Computational Linguistics}
}

@inproceedings{wu2024clasheval,
  title     = {{ClashEval}: Quantifying the Tug-of-War between an {LLM}'s Internal Prior and External Evidence},
  author    = {Wu, Kevin and Wu, Eric and Zou, James},
  booktitle = {Advances in Neural Information Processing Systems 37, Datasets and Benchmarks Track},
  year      = {2024}
}

@inproceedings{zou2025poisonedrag,
  title     = {{PoisonedRAG}: Knowledge Corruption Attacks to Retrieval-Augmented Generation of Large Language Models},
  author    = {Zou, Wei and Geng, Runpeng and Wang, Binghui and Jia, Jinyuan},
  booktitle = {34th USENIX Security Symposium},
  year      = {2025}
}

@inproceedings{ouyang2022,
  title     = {Training Language Models to Follow Instructions with Human Feedback},
  author    = {Ouyang, Long and Wu, Jeff and Jiang, Xu and Almeida, Diogo and Wainwright, Carroll L. and Mishkin, Pamela and Zhang, Chong and Agarwal, Sandhini and Slama, Katarina and Ray, Alex and Schulman, John and Hilton, Jacob and Kelton, Fraser and Miller, Luke and Simens, Maddie and Askell, Amanda and Welinder, Peter and Christiano, Paul and Leike, Jan and Lowe, Ryan},
  booktitle = {Advances in Neural Information Processing Systems 35},
  pages     = {27730--27744},
  year      = {2022}
}

@inproceedings{rafailov2023,
  title     = {Direct Preference Optimization: Your Language Model is Secretly a Reward Model},
  author    = {Rafailov, Rafael and Sharma, Archit and Mitchell, Eric and Ermon, Stefano and Manning, Christopher D. and Finn, Chelsea},
  booktitle = {Advances in Neural Information Processing Systems 36},
  year      = {2023}
}

@article{shumailov2024,
  title   = {{AI} Models Collapse when Trained on Recursively Generated Data},
  author  = {Shumailov, Ilia and Shumaylov, Zakhar and Zhao, Yiren and Papernot, Nicolas and Anderson, Ross and Gal, Yarin},
  journal = {Nature},
  volume  = {631},
  pages   = {755--759},
  year    = {2024}
}

@inproceedings{xiao2024sinks,
  title     = {Efficient Streaming Language Models with Attention Sinks},
  author    = {Xiao, Guangxuan and Tian, Yuandong and Chen, Beidi and Han, Song and Lewis, Mike},
  booktitle = {The Twelfth International Conference on Learning Representations},
  year      = {2024}
}
\bibliographystyle{nejlt_bib}

\clearpage
\onecolumn
\appendix

\section{Study 2: Second Annotator}
\label{app:s2second}
With the second annotator's scores and the same corrected analysis: H1, 97.2\% $\to$ 50.0\% ($n=106$; 51 vs.\ 1 discordant; $p=2.4\times10^{-14}$); H2, 1 of 51 correct in isolation; H3, 52.6\% vs.\ 57.7\% error-free ($n=97$; 13 vs.\ 18 discordant; $p=0.47$).

\clearpage
\section{Probe 1b Pilot}
\label{app:pilot}
The pilot used 24 facts (12 capitals, 6 chemical symbols, 6 currencies), each asked once without context and three times after a nine-sentence text with the false version placed first, in the middle or last. It was run on the 1.4B and 2.8B pairs and on Qwen2.5-1.5B. Context-followed counts on the 72 context items: Pythia-2.8B 72, Mamba-2.8B 72, Pythia-1.4B 70, Mamba-1.4B 17, Qwen2.5-1.5B 72. Prior accuracy without context was 20, 22, 20, 24 and 23 of 24. Because both 2.8B models were at ceiling, the main probe added harder facts, 31-line texts and support items.

\clearpage
\section{Example Items}
\paragraph{Probe 1b, conflict item.} Fact: capital of Australia (true answer Canberra, planted answer Sydney); 31-line text, planted line in the middle. The two worked examples with invented facts precede the item and are omitted here, as are lines 4--12 and 20--28 of the text.
\begin{quote}\small
\begin{verbatim}
Read the text and answer using ONLY what the text says.

Text:
- The sky appears blue during a clear day.
- Water freezes at zero degrees Celsius.
- A triangle has three sides.
  ...
- Snow is made of ice crystals.
- Cows produce milk.
- The capital of Australia is Sydney.
- A circle has no corners.
- Birds lay eggs.
  ...
- Bees can fly.
- Salt dissolves in water.
- Leaves fall from many trees in autumn.

Question: According to the text, what is the capital of Australia?
Answer:
\end{verbatim}
\end{quote}
A response containing ``Sydney'' and not ``Canberra'' counts as context-followed; the reverse counts as prior-fallback. The baseline item for the same fact is simply \texttt{Question: what is the capital of Australia? Answer:}.

\paragraph{Study 2.} Question: ``Who developed Japan's first arc lamp?'' (gold answer: Ichisuke Fujioka). GPT-OSS-120B names a different engineer and then adds an employer and a demonstration city. Condition B replaces the name with ``Ichisuke Fujioka''; condition D replaces it with ``Shoichi Tanaka'', which is plausible but wrong. In both, the model continues from the inserted name.

\end{document}